\documentclass[9pt,journal,letterpaper,twocolumn]{IEEEtran}
\usepackage{amssymb,amsmath}
\usepackage{balance}
\usepackage{epsfig}
\usepackage{algorithm} 
\usepackage{algorithmic} 
\usepackage{booktabs}
\usepackage{graphicx,subfigure}
\usepackage{graphics}
\usepackage{array}
\usepackage{multirow}
\usepackage{flushend}
\usepackage{color}
\usepackage{amsthm}

\usepackage{ragged2e}
\usepackage{multicol}

\makeatletter

\newcommand{\Rmnum}[1]{\expandafter\@slowromancap\romannumeral #1@}
\makeatother
\ifCLASSOPTIONcompsoc
\usepackage[nocompress]{cite}
\else
\usepackage{cite}
\fi

\ifCLASSINFOpdf
\else
\fi

\begin{document}
	%

	\title{Social Anchor-Unit Graph Regularized Tensor Completion for Large-Scale Image Retagging}

	%
	%
	%
	%
	
	\author{
		Jinhui~Tang,
		Xiangbo~Shu,
		Zechao~Li,
		Yu-Gang Jiang,
		and Qi~Tian,~\IEEEmembership{Fellow,~IEEE}
		\thanks{\em J. Tang, X. Shu, and Z. Li are with the School of Computer Science and Engineering, Nanjing
			University of Science and Technology, Nanjing 210094, China. E-mail:$\{$jinhuitang, shuxb, zechao.li$\}$@njust.edu.cn. (Corresponding author: Xiangbo Shu)}
		\thanks{\em Y.-G. Jiang is with the School of Computer Science, Fudan University, Shanghai 201203, China. E-mail: ygj@fudan.edu.cn}
				\thanks{\em Q. Tian is with the Department of Computer Science, University of Texas at San Antonio, San Antonio, TX 78249-1604, USA. E-mail: qi.tian@utsa.edu}}
	
	%
	%

\markboth{SUBMISSION~FOR~IEEE~TRANSACTIONS~ON~PATTERN~ANALYSIS~AND~MACHINE~INTELLIGENCE, 2018}%
{SUBMISSION~FOR~IEEE~TRANSACTIONS~ON~PATTERN~ANALYSIS~AND~MACHINE~INTELLIGENCE, 2018}
	%



	\IEEEcompsoctitleabstractindextext{
		\begin{abstract}
			\justifying
	Image retagging aims to improve the tag quality of social images by completing the missing tags, recrifying the noise-corrupted tags, and assigning new high-quality tags. Recent approaches simultaneously explore visual, user and tag information to improve the performance of image retagging by mining the tag-image-user associations. However, such methods will become computationally
	infeasible with the rapidly increasing number of images, tags and users. It has been proven that the anchor graph can significantly accelerate large-scale graph-based learning by exploring only a small number of anchor points. Inspired by this, we propose a novel Social anchor-Unit GrAph Regularized Tensor Completion (SUGAR-TC) method to efficiently refine the tags of social images, which is insensitive to the scale of data. First, we construct an anchor-unit graph across multiple domains (\textit{e.g.}, image and user domains) rather than traditional anchor graph in a single domain. Second, a tensor completion based on Social anchor-Unit GrAph Regularization (SUGAR) is implemented to refine the tags of the anchor images. Finally, we efficiently assign tags to non-anchor images by leveraging the relationship between the non-anchor units and the anchor units. Experimental
	results on a real-world social image database well demonstrate the effectiveness and efficiency of SUGAR-TC, outperforming the state-of-the-art methods.
			\justifying
		\end{abstract}
		
		\begin{IEEEkeywords}
			Image retagging, anchor graph, tensor completion, image retrieval.
		\end{IEEEkeywords}}

		\maketitle

		\IEEEdisplaynontitleabstractindextext

		%
		\IEEEpeerreviewmaketitle

\section{Introduction}


In the past decades, various social image retagging methods~\cite{zhu2010image,sang2011exploiting,wu2013tag,tang2017tri,sang2012user,DBLP:journals/pami/QiATJH12} have been proposed to improve the tag quality of social images. At the early stage, image retagging methods refine tags of social images by utilizing semantic correlation among tags and visual similarity among images~\cite{liu2010image,yang2011mining}. A basic assumption of such methods is that two images with high visual similarity should have similar semantic tags, and vice versa. To well leverage the available image-tag inter-association, some researchers improved the matrix completion based on the low-rank decomposition to simultaneously recover the missing tags, and remove the noisy tags~\cite{xu2017non,li2016locality}. Candes and Plan~\cite{candes2010matrix} have proven that the matrix completion model enables to complete the missing entries from a small number of observed entries in the original matrix between the dyadic data. However, {the aforementioned methods achieve unsatisfied retagging results when there exists disambiguation between the visual content and label taxonomy (\textit{\textit{e.g.}}, WordNet taxonomy~\cite{lin1997using})}.

\begin{figure}[!t]
	\small{
		\centering
		\subfigure[Traditional methods~\cite{sang2012user,tang2017tri}]
		{
			\includegraphics[scale=0.4]{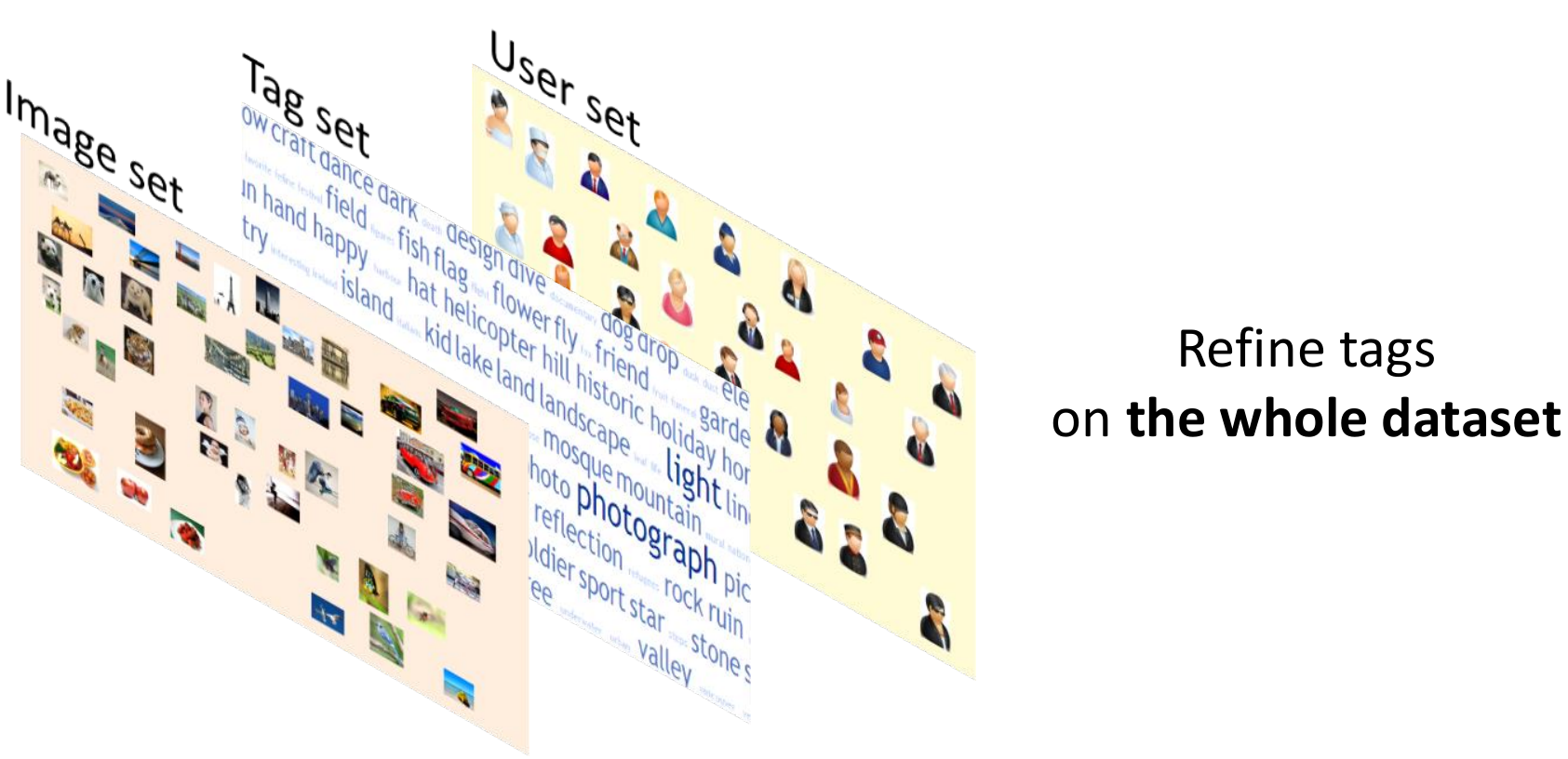}
		}
		\subfigure[The proposed SUGAR-TC]
		{
			\includegraphics[scale=0.38]{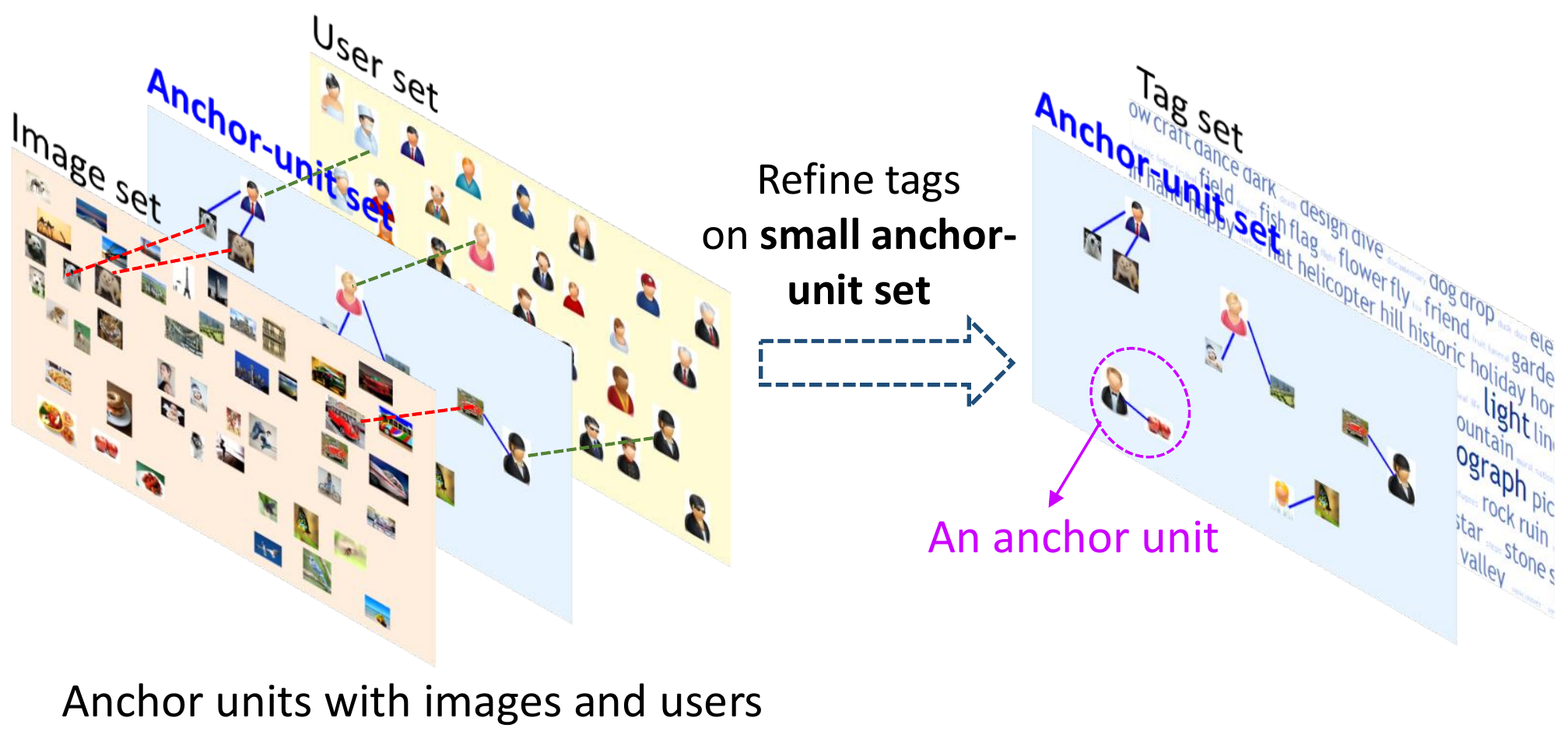}
		}
		\vspace{-2mm}
		\caption{Conventional methods refine tags of images by exploring the inter-association among all data. This comes at a price of increased computational cost when faced with a larger number of images. The proposed method refines tag on a smaller-scale anchor-unit set, and then assigns tags to all non-anchor images in an efficient way, regardless of the data scale.}
		\vspace{-2mm}
		\label{fig_idea}
	}
\end{figure}

\begin{figure*}[t]
	\centering
	\includegraphics[scale=0.425]{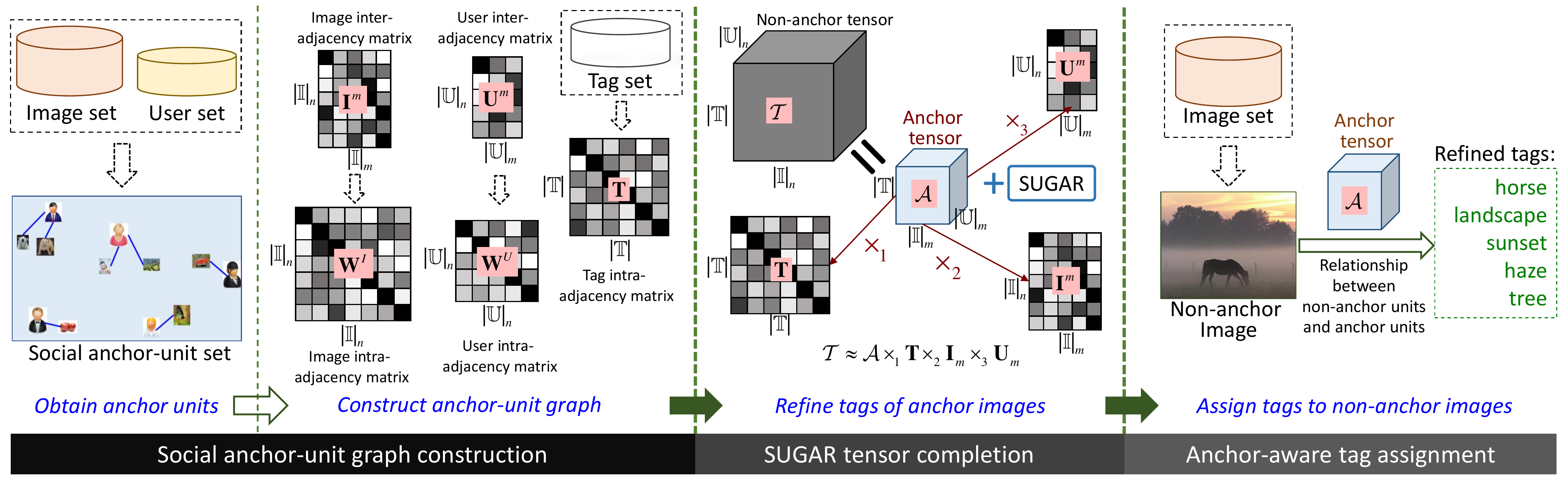}
	\vspace{-6mm}
	\caption{The whole framework of SUGAR-TC for social image retagging. (a) Social anchor-unit graph construction: obtain anchor units and construct anchor-unit graph with the inter- and intra- adjacency edges. (b) SUGAR tensor completion: refine tags of anchor images. (c) Anchor-aware tag assignment: assign tags to non-anchor images via the inter-association between non-anchor units and anchor units.}
	\vspace{-2mm}
	\label{fig2}
\end{figure*}

Intuitively, a set of images uploaded by one common user tends to have close relations. That is to say, the user information (\textit{\textit{e.g.}}, user interests and backgrounds) bridges the inter-relationship between tags and social images~\cite{cui2014social}, especially for some certain tags, \textit{\textit{e.g.}}, geo-related tags, event tags, \textit{etc.} Therefore, several methods were proposed to simultaneously leverage the visual information, tag information, and user information for social image retagging. Sang \textit{et al.}~\cite{sang2011exploiting} attempted to model the inter-association among users, images and tags, and presented a tensor completion based on the low-rank decomposition to refine the social image tags. However, since the process of tensor completion generates many large-scale temporary matrices and tensors, it requires extremely large computational cost. 
To handle this problem, Tang \textit{et al.}~\cite{tang2017tri} proposed Tri-clustered Tensor Completion (TTC) to divide the original tag-image-user tensor among all tags, images, and users into several sub-tensors, and implement the sub-tensor completion in a parallel way. However, the computational cost still increases accompanied with the number of images increasing.

To this end, we aim to develop an efficient image retagging framework to assign high-quality tags to social images, regardless of the data scale. We investigate that anchor graph can significantly accelerate large-scale graph-based learning by only exploring a small number of anchor points on the graph~\cite{liu2011hashing,wang2017learning}. Inspired by this, we propose a {Social anchor-Unit GrAph Regularization (SUGAR)} model on a tag-image-user graph by exploring a small number of representative anchor units, where one anchor unit consists of an anchor image and an anchor user, as shown in Figure~\ref{fig_idea}. Accordingly, we propose a novel {Social anchor-Unit GrAph Regularized Tensor Completion (SUGAR-TC)} method to efficiently refine the tags of social images. The framework of SUGAR-TC is shown in Figure 2, which mainly consists of three modules. (a) {Anchor-unit graph construction module.} We employ the co-clustering algorithm to obtain the representative anchor units in multiple domains (\textit{i.e.}, image and user domains) rather than the traditional anchors in a single domain, and then construct a social anchor-unit graph. (b) {SUGAR tensor completion module.} We propose a tensor completion model with Social anchor-Unit GrAph Regularization (SUGAR). It completes the anchor tensor by decomposing the non-anchor tensor to refine the tags of anchor images. (c) Anchor-aware tag assignment module. By leveraging the potential relationships between non-anchor units and anchor units, we use the weighted average tags of anchor images to efficiently assign high-quality tags to the non-anchor images.

Overall, the main contributions of this work are two-fold. (1) We propose an efficient Social anchor-Unit GrAph Regularized Tensor Completion (SUGAR-TC) method for social image retagging, even when the data scale is very large. (2) To the best of our knowledge, it is the first time that social anchor unit
across multiple domains rather than the traditional anchors in a single domain is  presented.  In the experiments, the proposed SUGAR-TC gains superior performance on both effectiveness and efficiency compared with the state-of-the-art methods.

\section{Related Work}
\subsection{Social Image Retagging}
The goal of social image retagging is to improve tag quality by recovering missing tags and removing noisy tags~\cite{wang2012assistive,li2016socializing}. It is an essential step for large-scale tag-based image retrieval. The image retagging task is closely related to tag refinement~\cite{fu2015image,tang2017tri}, tag completion~\cite{wu2013tag}.   
Overall, social image retagging methods have gone through two stages, \textit{i.e.}, image-tag association based image retagging~\cite{xu2009tag,yang2011mining,chen2013fast,liu2010image,zhu2010image,wu2013tag,li2018deep,liu2012image,li2017weakly,lin2013image,li2014image}, and tag-image-user association based image retagging~\cite{sang2011exploiting,tang2017tri,rafailidis2014content,sang2012user}.

The image-tag association based methods focus on exploring inter- and intra- associations among images and tags. As one of the most classic works,  Liu~\textit{et al.}~\cite{liu2010image} assumed that two images with higher visual similarity are more likely to have similar tags, or common tags. Yang \textit{et al.}~\cite{yang2011mining} proposed to mine the multi-tag intra-association for tag expansion and denoising.
Chen \textit{et al.}~\cite{chen2013fast} proposed co-regularized learning with two classifiers by jointly mapping visual features and text features into a common subspace. 

Matrix completion based on the low-rank decomposition is popularly applied for image retagging. As an early study, Zhu \textit{et al.}~\cite{zhu2010image} solved the image retagging problem by factorizing  the image-tag inter-association matrix into an approximately low-rank completed matrix and a sparse error matrix, where this low-rank completed matrix reveals the image-tag associations. Besides, Feng \textit{et al.}~\cite{feng2014image} theoretically analyzed that the matrix completion method is able to simultaneously recover missing tags and remove noisy tags even with a limited number of observations. Motivated by the
success of low-rank model in image tagging,  Xu
{\em et al.}~\cite{xu2017non} proposed Non-linear Matrix Completion (NMC) for social
image tagging by constructing the original image-tag matrix with a
non-linear kernel mapping. Li \textit{\em et al.}~\cite{li2016locality} divided
the original image-tag matrix into several sub-matrices, and proposed
a Locality Sensitive Low-Rank (LSLR) model on each sub-matrix to
recover the image-tag matrix via matrix factorization.

Tag-image-user association based image retagging methods construct a tag-image-user inter-association tensor instead of the image-tag inter-association matrix to explore the important user information~\cite{sang2011exploiting}. It can further improve the performance of image retagging with the help of user information. Tang \textit{et al.}~\cite{tang2017tri} proposed a Tri-clustered Tensor Completion (TTC) framework to first divide the original super-sparse tensor into several sub-tensors, and complete all sub-tensors regularized by a tensor kernel. However, to a certain degree, TTC will break the tag-image-user inter-association when dividing the original super-sparse tensor into several sub-tensors.

\subsection{Anchor Graph-based Learning}
Graph-based learning methods have achieved impressive performance in various applications~\cite{deng2013visual}. 
However, such methods require much computational cost when the size of data rapidly increases. Therefore, several strategies have been proposed to  reduce the computational cost of graph-based
learning. To sum up, these strategies can be summarized into three categories. The first strategy uses neighborhood propagation from the approximate neighborhood graph to current graph~\cite{harwood2016fanng,suzuki2017centered,tang2011image,chen2012personalized}. The second strategy utilizes hashing to improve the performance in terms of efficiency~\cite{song2013effective,norouzi2014fast,liu2014discrete,jiang2015scalable}. The last strategy employs the anchor graph to simultaneously reduce the computational cost and memory cost~\cite{liu2010large,wang2016scalable,kim2013multi,wang2017learning,liu2012robust}. On an anchor graph, anchors covering vast data point cloud can predict the label for each data point, even when the data size is large. As a result, the anchor graph model has been successfully applied to solve many practical tasks, including image retrieval~\cite{xu2015emr}, face recognition~\cite{xiong2013face}, image classification~\cite{wang2017learning}, object tracking~\cite{wu2015robust}, and so on. In this work, we extend the anchor graph 
in a single domain to the anchor-unit graph across multiple domains, to simultaneously explore the association information among different domains.

\section{The Proposed Framework}
This section introduces the whole framework of SUGAR-TC, as shown in Figure~\ref{fig2}. It includes three main modules, {\em i.e.}, Social anchor-unit
graph construction module, SUGAR tensor completion module, and Anchor-aware
tag assignment module. In this paper, tensors, matrices, vectors, variables and
sets are denoted by calligraphic uppercase letters (\textit{\textit{e.g.}}, $\mathcal{A}$), bold uppercase letters (\textit{\textit{e.g.}}, ${\bf T}$), bold lowercase
letters (\textit{\textit{e.g.}}, ${\bf d}$), letters (\textit{\textit{e.g.}}, m, I) and
blackboard bold letters (\textit{\textit{e.g.}}, $\mathbb{I}$), respectively. ${\bf T}_{i,j}$ denotes the $i$-th row and $j$-th column entry of ${\bf T}$. $\mathcal{T}_{i,j,k}$ denotes the $i$-th row, $j$-th column and $k$-th tube entry of $\mathcal{T}$. $|\mathbb{I}|$ denotes the number of data in $\mathbb{I}$. ${\mathcal A}_{[1]}^{}$, ${\mathcal A}_{[2]}^{}$ and ${\mathcal A}_{[3]}^{}$ are the matrix form of ${\mathcal A}$ by accumulating the entries of ${\mathcal A}$ along the row, column, and tube axes, respectively. For  convenience,  some important
notations are defined in Table~\ref{tab_notation}.

\subsection{Background and Problem Definition}
In this work, three types of data are collected from photo sharing websites, namely the image set
$\mathbb{I}=\{x_i\}_{i=1}^{|\mathbb{I}|}$, the tag set $\mathbb{T}=\{t_j\}_{j=1}^{|\mathbb{T}|}$ and the user set $\mathbb{U}=\{u_k\}_{k=1}^{|\mathbb{U}|}$, where $x_i, t_j$ and $u_k$ denote the \textit{i}-th image, the \textit{j}-th tag, and the \textit{k}-th user, respectively.  Since many tags of images freely given by users with various backgrounds and interests are ambiguous, noisy and incomplete, the original user-provided tags are weakly-supervised, which cannot represent the semantics of images well. If we directly implement the tag-based image retrieval on social images with original tags, the performance of retrieval will degrade. In this work, we aim to refine the tags of social images by utilizing the associations among tags, images and users. However, the numbers of images, tags and users are large, and even huge on photo sharing websites. This will dramatically increase the computational cost.

   It has been proven that the anchor graph can significantly accelerate large-scale graph-based learning by exploring only a small number of anchor points. However, traditional anchor graph is constructed only in image domain. To leverage the user information, we extend the traditional anchor to anchor unit, which is a tuple including an anchor image and an anchor user. And the graph constructed on the anchor units is called anchor-unit graph.

 Let ${\mathbb I}_n$, ${\mathbb U}_n$, ${\mathbb I}_m$, and ${\mathbb U}_m$ indicate the non-anchor image set, the non-anchor user set, the anchor image set, and the anchor user set respectively, where $|{\mathbb I}_n|+|{\mathbb I}_m|=|\mathbb{I}|$ and $|{\mathbb U}_n|+|{\mathbb U}_m|=|\mathbb{U}|$, we construct a tag-image-user non-anchor tensor $\mathcal{T}\in\Bbb{R}^{|\Bbb{T}|\times|\Bbb{I}_n|\times|\Bbb{U}_n|}$ among all the tags, non-anchor images and non-anchor users, and a tag-image-user anchor tensor $\mathcal{A}_0\in\Bbb{R}^{|\Bbb{T}|\times|\Bbb{I}_m|\times|\Bbb{U}_m|}$ among all the tags, anchor images and anchor users, in which some associations are incorrect or missing. Let ${\mathcal{A}}\in\Bbb{R}^{|\Bbb{T}|\times|\Bbb{I}_m|\times|\Bbb{U}_m|}$ denote the refined anchor tensor among the refined tags, anchor images and anchor users, this work first aims to learn ${\mathcal{A}}$ with the help of the inter-association in $\mathcal{T}$ and $\mathcal{A}_0$. After that, benefiting from the tag-image-user association in ${\mathcal{A}}$, we can efficiently retag the non-anchor images based on the inter-relationship between non-anchor units and anchor units on the anchor-unit graph.

%


\subsection{Social Anchor-Unit Graph Construction}
To construct the anchor-unit graph, we need to obtain the anchor units from the image and user data. Generally speaking, the perfect anchor units should satisfy two conditions: 1) they can adequately represent the distribution of image and use data; 2) the number of the anchor units should be much smaller than the number of images~\cite{liu2010large}. Existing methods employ the K-means
clustering to obtain the traditional anchors~\cite{liu2014discrete,wang2016scalable}. However, K-means algorithm performing in the single domain cannot cluster the image and user data simultaneously. Fortunately, co-clustering algorithm can construct the association between two types of data by an associated matrix, and then cluster rows and columns of this matrix simultaneously into several co-clusters~\cite{dhillon2001co}. Therefore, we can build an associated image-user matrix ${\mathcal D}_{[1]}\in {\mathbb{R}^{|\mathbb{I}|\times |\mathbb{U}|}}$ by accumulating entries of tag-image-user tensor $\mathcal {D} $ along the tag axis, and adopt the co-clustering algorithm instead of the K-means algorithm to find $C$ co-cluster centers. Subsequently, we select $m_c$ image-user units (a user uploads this image) that is most close to the $c$-th co-cluster center as the anchor unit. In total, we can obtain $m$ image-user anchor units for $C$ co-cluster centers, where $m=m_c \times C$.

\begin{table}[t!]
	\renewcommand{\arraystretch}{1.2}
	\centering
	\caption{Notations and definitions.}
	\label{tab_notation}
	{\scriptsize
		\begin{tabular}{ll}
			\hline
			\hspace{-0.5em}Notation & \hspace{-0.5em}Definition\\
			\hline
			\hspace{-0.5em}$\mathbb{I}$& \hspace{-0.5em}Image set.\\
			\hspace{-0.5em}$\mathbb{U}$& \hspace{-0.5em}User set. \\
			\hspace{-0.5em}$\mathbb{T}$& \hspace{-0.5em}Tag set. \\
			\hspace{-0.5em}$\mathbb{I}_n$& \hspace{-0.5em}Non-anchor image set.\\
			\hspace{-0.5em}$\mathbb{U}_n$& \hspace{-0.5em}Non-anchor user set. \\
			\hspace{-0.5em}$\mathbb{I}_m$& \hspace{-0.5em}Anchor image set.\\
			\hspace{-0.5em}$\mathbb{U}_m$& \hspace{-0.5em}Anchor user set.\\
			\hspace{-0.5em}$\mathcal{D}$& \hspace{-0.5em}Original tag-image-user tensor on $\mathbb{T}$, $\mathbb{I}$ and $\mathbb{U}$.\\
			\hspace{-0.5em}$\mathcal{T}$& \hspace{-0.5em}Original tag-image-user non-anchor tensor on $\mathbb{T}$, $\mathbb{I}_n$ and $\mathbb{U}_n$.\\
			\hspace{-0.5em}$\mathcal{A}_0$& \hspace{-0.5em}Original tag-image-user anchor tensor on $\mathbb{T}$, $\mathbb{I}_m$ and $\mathbb{U}_m$.\\
			\hspace{-0.5em}$\mathcal{A}$&\hspace{-0.5em}Refined tag-image-user anchor tensor. ({\em to be learned})\\
			\hspace{-0.5em}${\bf I}_m$& \hspace{-0.5em}Inter-adjacency matrix between non-anchor image and anchor image. \\
			\hspace{-0.5em}${\bf U}_m$& \hspace{-0.5em}Inter-adjacency matrix between non-anchor user and anchor user. \\
			\hspace{-0.5em}${\bf W}_I$& \hspace{-0.5em}Intra-adjacency matrix between non-anchor image and non-anchor image. \\
			\hspace{-0.5em}${\bf W}_U$& \hspace{-0.5em}Intra-adjacency matrix between non-anchor user and non-anchor user. \\
			\hspace{-0.5em}${\bf T}$& \hspace{-0.5em}Intra-adjacency matrix between tag and tag. \\
			\hline
		\end{tabular}
	}
\end{table}

Based on the $m$ anchor units, we construct a social anchor-unit graph $\Omega \{{\mathbb I}_n, {\mathbb U}_n, {\mathbb I}_m, {\mathbb U}_m, {\mathbb T}, \omega\}$, where $\omega$ indicates the collection of the adjacency edges between the data points. To measure the weight of each edge in anchor-unit graph $\Omega$, we design two types of inter-adjacency matrices between non-anchor data and anchor data (\textit{i.e.}, image inter-adjacency matrix and user inter-adjacency matrix), and three types of intra-adjacency matrices among non-anchor images, non-anchor users and tags (\textit{i.e.}, image intra-adjacency matrix, user intra-adjacency matrix, and tag intra-adjacency matrix).

{\bf Inter-adjacency matrices.} First, we design the image inter-adjacency matrix ${{\bf{I}}^m}\in{\mathbb R} ^{{|{\mathbb I}_n|} \times {|{\mathbb I}_m|}}$ between $|{\mathbb I}_n|$ non-anchor images and ${|{\mathbb I}_m|}$ anchor images, \textit{i.e.},
\begin{equation}
	\label{eq1}
	{\bf{I}}^m_{i,j}=exp
	\left(
	-\dfrac{||{\bf d}_{x_i}-{\bf d}_{x_j}||_2^2}{\sigma^2}
	\right),
	\vspace{-0mm}
\end{equation}
where $\sigma$ is a parameter of the RBF kernel, ${\bf d}_{x_i}$ and ${\bf d}_{x_j}$ indicate features (\textit{i.e.}, CNN feature~\cite{krizhevsky2012imagenet}) of a non-anchor image $x_i$ and an anchor image $x_j$, respectively. Here, if ${\bf{I}}^m_{i,j} < 10^{-4}$, we set ${\bf{I}}^m_{i,j} = 0$ to obtain a sparse matrix ${{\bf{I}}^m}$ for reducing the memory cost. 

Second, we also design user inter-adjacency matrix ${{\bf{U}}^m}\in{\mathbb R} ^{{|{\mathbb U}_n|} \times {|{\mathbb U}_m|}}$ between the $|{\mathbb U}_n|$ non-anchor users and $|{\mathbb U}_m|$ anchor users. Similar to \cite{tang2017tri}, we assume that two users with higher co-occurrence are more likely to be related with each other, and vice versa, namely
\begin{equation}
	\label{eq2}
	{\bf{U}}^m_{i,j}=
	\dfrac{{N}(u_i,u_j)}{{N}(u_i)+{N}(u_j)-{N}(u_i,u_j)}
	,
\end{equation}
where ${N}(u_i,u_j)$ denotes the number of groups that both a non-anchor user $u_i$ and an anchor user $u_j$ join, and ${N}(u_i)$ is the number of groups that a non-anchor user $u_i$ joins, ${N}(u_j)$ is the number of groups that an anchor user $u_j$ joins.

{\bf Intra-adjacency matrices.} We first design image intra-adjacency matrix ${{\bf{W}}^I}\in{\mathbb R} ^{{|{\mathbb I}_n|} \times {|{\mathbb I}_n|}}$ among the non-anchor images. One strategy is that we can measure the association between two non-anchor images by Eq.~\eqref{eq1}. However, the number of  non-anchor images is much larger than the number of anchor images, which will increase the computational cost. Alternately, since anchor images are very representative in the whole image set, we can measure the association between non-anchor images by exploring links among common anchor images, which has also been claimed in~\cite{wang2017learning,liu2011hashing}. For two non-anchor images, the more number of common anchor images the two non-anchor images share, the stronger association such two non-anchor images have. Thus, ${{\bf{W}}^I}\in{\mathbb R} ^{{|{\mathbb I}_n|} \times {|{\mathbb I}_n|}}$ can be computed as follows,
\begin{equation}
	\label{eq3}
	{\bf{W}}_{}^I = {{\bf{I}}^m} {\left( {{\Lambda ^I}} \right)^{ - 1}} {\left( {{{\bf{I}}^m}} \right)^T},
\end{equation}
where the diagonal matrix $\Lambda^I$ is defined as  $\Lambda _{j,j}^I{\rm{ = }}\sum\limits_{i = 1}^{|{\mathbb I}_n|} {{\bf I}_{i,j}^m}$ ($1\leqslant j\leqslant  |{\mathbb I}_m| $). From Eq~\eqref{eq3}, if ${}{\bf{W}}_{i,j}^I>0$ , it means two non-anchor images share at least one anchor user, and otherwise ${\bf{W}}^I_{i,j}=0$.  

Second, we use the same idea to design user intra-adjacency matrix ${{\bf{W}}^U}\in{\mathbb R} ^{{|{\mathbb U}_n|} \times {|{\mathbb U}_n|}}$ among the non-anchor users, as follows,
\begin{equation}
	\label{eq4}
	{\bf{W}}_{}^U = {{\bf{U}}^m} {\left( {{\Lambda ^U}} \right)^{ - 1}} {\left( {{{\bf{U}}^m}} \right)^T},
\end{equation}
where the diagonal matrix $\Lambda^U$ is defined as  $\Lambda_{j,j}^U{\rm{ = }}\sum\limits_{i = 1}^{|{\mathbb U}_n|} {{\bf U}_{i,j}^m}$. 

Following the definitions in~\cite{sang2012user,tang2017tri}, we design the tag intra-adjacency matrix ${\bf T}\in{\mathbb R} ^{{|{\mathbb T}|} \times {|{\mathbb T}|}}$ by using the categorical relations and the co-occurrence, \textit{i.e.}
\begin{equation}
	\label{eq4.5}
	\begin{split}
		{\bf T}_{i,j}=a_1 \dfrac{N(t_i,t_j)}{N(t_i)+N(t_j)-N(t_i,t_j)}
		+a_2 \dfrac{2\!\cdot \!C (L(t_i,t_j))}{C (t_i)+C (t_j)},
	\end{split}
	\vspace{-0mm}
\end{equation}
where $a_1$ and $a_2$ denote the weight coefficients ($a_1 +a_2=1$); $N(t_i)$ denotes the occurrence count of tag $t_i$ in a dataset; $N(t_i,t_j)$ denotes the co-occurrence count for tags $t_i$ and $t_j$  in a dataset; $C({t_i}) =  - log\left( {p\left( {{t_i}} \right)} \right)$ is the information content of tag $t_i$; $p(t_i)$ is the probability of tag $t_i$; and $L(t_i,t_j)$ is the least common sub-sumer of tags $t_i$ and $t_j$ in the WordNet taxonomy~\cite{lin1997using}. The least common sub-sumer of two tags in WordNet is the sumer that does not have any children that are also the sub-sumer of two tags.

\subsection{SUGAR Tensor Completion}

It has been proved that anchor graph can efficiently deal with the standard multi-class semi-supervised learning problem~\cite{liu2010large}. Motivated by this, we extend the existing Anchor Graph Regularization~\cite{wang2016scalable} to a novel Social anchor-Unit GrAph Regularization (SUGAR). Specifically, it can be assumed that two images linked by the same anchor image should have closely similar tags and closely related users, two users linked by the same anchor user prefer to upload images with closely similar tags.
Then, we can define the regularization as follows,
\begin{equation}
	\label{eq6.0}
	\begin{aligned}
		{{{\Theta}_1}}=& \frac{\lambda_1}{2}\sum\limits_{i = 1,j = 1}^{|{\mathbb I}_n|} {{\bf{W}}_{i,j}^I {||
				{\mathcal A} \times_2 {\bf{I}}_{i,:}^m
				- {\mathcal A} \times_2 {\bf{I}}_{j,:}^m
				||_F^2} }\\
		&+\frac{\lambda_2}{2}\sum\limits_{i = 1,j = 1}^{|{\mathbb U}_n|} {{\bf{W}}_{i,j}^U {||
				{\mathcal A} \times_3 {\bf{U}}_{i,:}^m
				- {\mathcal A} \times_3 {\bf{U}}_{j,:}^m
				||_F^2}},
	\end{aligned}
\end{equation}
where $\lambda_1$ and $\lambda_2$ are nonnegative coefficients to control the penalty of the corresponding regularization. The matrix ${\mathcal A}{ \times _2}{\bf{I}}^m_{i,:}$ denotes the $2$-mode product\footnote{More details about $n$-mode product can be found in~\cite{kolda2009tensor}.} of ${\mathcal A}$ and ${\bf{I}}^m_{i,:}$, and describes the inter-association between the tags and the anchor users corresponding to the $i$-th non-anchor image. The matrix ${\mathcal A}{ \times _3}{\bf{U}}^m_{i,:}$ describes the inter-association between the tags and the anchor images corresponding to the $i$-th non-anchor user. 

\begin{algorithm}[tb]
	\renewcommand{\algorithmicrequire}{\textbf{Input:}}
	\renewcommand\algorithmicensure {\textbf{Output:} }
	\caption{ SUGAR Tensor Completion} 
	\small
	\label{alg1}
	\begin{algorithmic} [1]
		\REQUIRE
		~\\
		Original non-anchor tensor $\mathcal{T}$, original anchor tensor $\mathcal{A}_0$ on the anchor-unit set, adjacency matrices ${\bf T}$, ${\bf I}^m$, ${\bf U}^m$, ${\bf W}^I$ and ${\bf W}^U$, parameters $\alpha$, $\beta$, $\lambda_1$ and $\lambda_2$.
		\ENSURE
		~\\
		Completed anchor tensor ${\mathcal{A}}$.
		\renewcommand{\algorithmicrequire}{\textbf{Initialization:}}
		\REQUIRE $\mathcal{A}$.		
		\REPEAT
		\STATE Update ${\cal A}$ by Eq.~\eqref{eq12.1}.
		\UNTIL{Convergence.}
		\RETURN ${\mathcal{A}}$.
	\end{algorithmic}
\end{algorithm}

The great number of original tags provided by users on websites involve the important supervised information, which can guide us to mine the inter-association among images, tags, and users~\cite{sang2011exploiting}. 
 Accordingly, we construct the tag-image-user non-anchor tensor $\mathcal{T}\in\Bbb{R}^{|\Bbb{T}|\times|\Bbb{I}_n|\times|\Bbb{U}_n|}$ among tags, non-anchor images and non-anchor users. Specifically, if the \textit{i}-th non-anchor image uploaded by the \textit{k}-th non-anchor user is annotated with the \textit{j}-th tag, we set $\mathcal{T}_{i,j,k}=1$, otherwise $\mathcal{T}_{i,j,k}=0$, where $1\leqslant i\leqslant|\Bbb{T}|$, $1\leqslant j\leqslant|\Bbb{I}_n|$, and $1\leqslant k\leqslant|\Bbb{U}_n|$. For a tensor with a few available entries, the tensor completion algorithm can estimate missing
entries and remove the noisy ones by reconstructing an approximately low-rank tensor $\tilde{\mathcal{T}}$, as follows,
\begin{equation}
	\label{eq6}
	\min_{\tilde {\mathcal{T}}} {||\mathcal{T}-\tilde{\mathcal{T}}||_F^2}.
	\vspace{-0mm}
\end{equation}

The Tucker decomposition~\cite{kolda2009tensor} of tensor provides a factorization way to solve the low-rank tensor $\tilde{\mathcal{T}}$ by the following objective function, \textit{i.e.},
\begin{equation}
	\label{eq7}
	\min_{\tilde {\mathcal{T}}} {||\mathcal{T}-{{\mathcal S}{ \times _1}{\bf{B}}{ \times _2}{\bf{C}}{ \times _3}{\bf{D}}}||_F^2},
	\vspace{-0mm}
\end{equation}
where ${\tilde {\mathcal{T}}}={{\mathcal S}{ \times _1}{\bf{B}}{ \times _2}{\bf{C}}{ \times _3}{\bf{D}}}$. Here, ${\mathcal S}$ denotes the core tensor,  ${\bf B}$, ${\bf{C}}$, and ${\bf{D}}$ denote the factor matrices. Since ${{\bf{T}}}$, ${{\bf{I}}^m}$ and ${{\bf{U}}^m}$ describe the associations between tags and tags, non-anchor images and anchor images, as well as non-anchor users and anchor users, respectively, by setting ${\bf B}={\bf T}$, ${\bf C}={\bf{I}}^m$ and ${\bf D}={\bf{U}}^m$, the learned core tensor ${\mathcal {S}}$ can be regarded as the refined tensor ${\mathcal {A}}$. To well leverage the weakly supervised tag-image-user association, we introduce a new regularization term $\left\| {{\cal A} - {{\cal A}_0}} \right\|_F^2$, which  constrains the learned ${\mathcal {A}}$ to be consistent with the original tag-image-user anchor tensor ${\mathcal {A}}_0$. Finally, we can obtain the refined tensor $\mathcal {A}$ by minimizing the following tensor completion function $\Theta _2$, \textit{i.e.},
\begin{equation}
	\label{eq10}
	{{{\Theta }_2}}
	\!= \! \left\| {{\mathcal T} - {\mathcal A}{ \times _1}{\bf{T}}{ \times _2}{\bf{I}}^m{ \times _3}{\bf{U}}^m} \right\|_F^2  +\alpha\!\left\| {\mathcal A}-{\mathcal A}_0 \right\|_F^2+ \beta \left\| {\mathcal A} \right\|_F^2,
\end{equation}
where $\alpha$ and $\beta$ are nonnegative parameters to control the penalty of the regularization.
In Eq.~\ref{eq10}, the learned ${\mathcal A}$ is a low-rank and compact tensor, which can reveal the association among anchor-images, anchor-users and tags.

We integrate social anchor-graph regularization (\textit{i.e.}, Eq.~\eqref{eq6.0}) into Eq.~\ref{eq10}, and then obtain the completed anchor tensor ${\mathcal A}$ by minimizing an objective function $\Theta= {{\Theta}_1} +{{{\Theta}_2}}$, as follows,
\begin{equation}
	\label{eq10.1}
	\begin{aligned}
		\mathop {\min }\limits_{\mathcal A} &~\Theta  \\
		=  \mathop {\min }\limits_{\mathcal A} & \left\| {{\mathcal T} - {\mathcal A}{ \times _1}{\bf{T}}{ \times _2}{\bf{I}}^m{ \times _3}{\bf{U}}^m} \right\|_F^2 +\alpha\left\| {\mathcal A}-{\mathcal A}_0 \right\|_F^2+ \beta \left\| {\mathcal A} \right\|_F^2\\
		& +\frac{\lambda_1}{2}\sum\limits_{i = 1,j = 1}^{|{\mathbb I}_n|} {{\bf{W}}_{i,j}^I {||
				{\mathcal A} \times_2 {\bf{I}}_{i,:}^m
				- {\mathcal A} \times_2 {\bf{I}}_{j,:}^m
				||_F^2} }\\
		&+\frac{\lambda_2}{2}\sum\limits_{i = 1,j = 1}^{|{\mathbb U}_n|} {{\bf{W}}_{i,j}^U {||
				{\mathcal A} \times_3 {\bf{U}}_{i,:}^m
				- {\mathcal A} \times_3 {\bf{U}}_{j,:}^m
				||_F^2} }.
	\end{aligned}
\end{equation}

In this work, we learn ${\mathcal A}$ in an iteratively updating way. Specifically, we compute the partial derivatives of the objective function  $\Theta$ with respect to ${\mathcal A}$, \textit{i.e.},
\begin{equation}
	\begin{aligned}
		\frac{{\partial \Theta }}{{\partial {\mathcal A}}} =& 2{\mathcal A}{ \times _1}\left( {{{\bf{T}}^T}{\bf{T}}} \right){ \times _2}\left( {{{\left( {{{\bf{I}}^{\bf{m}}}} \right)}^T}{{\bf{I}}^{\bf{m}}}} \right){ \times _3}\left( {{{\left( {{{\bf{U}}^{\bf{m}}}} \right)}^T}{{\bf{U}}^{\bf{m}}}} \right)\\
		&- 2{\mathcal T}{ \times _1}{{\bf{T}}^T}{ \times _2}{\left( {{{\bf{I}}^{\bf{m}}}} \right)^T}{ \times _3}{\left( {{{\bf{U}}^{\bf{m}}}} \right)^T} + 2\alpha\left( {\mathcal A}-{\mathcal A}_0\right)+ 2\beta {\mathcal A} \\
		&+ {\lambda _1}\sum\limits_{i = 1,j = 1}^{|{\mathbb I}_n|} {{\bf{W}}_{i,j}^I\left( {{\mathcal A}{ \times _2}\left( {{{\bf{a}}_i}^T{{\bf{a}}_i}} \right) - {\mathcal A}{ \times _2}\left( {{{\bf{a}}_i}^T{{\bf{a}}_j}} \right)} \right)} \\
		&+ {\lambda _1}\sum\limits_{i = 1,j = 1}^{|{\mathbb I}_n|} {{\bf{W}}_{i,j}^I\left( {{\mathcal A}{ \times _2}\left( {{{\bf{a}}_j}^T{{\bf{a}}_j}} \right) - {\mathcal A}{ \times _2}\left( {{{\bf{a}}_j}^T{{\bf{a}}_i}} \right)} \right)} \\
		&+ {\lambda _2}\sum\limits_{i = 1,j = 1}^{|{\mathbb U}_n|} {{\bf{W}}_{i,j}^U\left( {{\mathcal A}{ \times _3}\left( {{{\bf{b}}_i}^T{{\bf{b}}_i}} \right) - {\mathcal A}{ \times _3}\left( {{{ {{{\bf{b}}_i}} }^T}{{\bf{b}}_j}} \right)} \right)}  \\
		&+ {\lambda _2}\sum\limits_{i = 1,j = 1}^{|{\mathbb U}_n|} {{\bf{W}}_{i,j}^U\left( {{\mathcal A}{ \times _3}\left( {{{\bf{b}}_j}^T{{\bf{b}}_j}} \right) - {\mathcal A}{ \times _3}\left( {{{ {{{\bf{b}}_j}} }^T}{{\bf{b}}_i}} \right)} \right)}\\
		=& 2{\mathcal A}{ \times _1}\left( {{{\bf{T}}^T}{\bf{T}}} \right){ \times _2}\left( {{{\left( {{{\bf{I}}^{\bf{m}}}} \right)}^T}{{\bf{I}}^{\bf{m}}}} \right){ \times _3}\left( {{{\left( {{{\bf{U}}^{\bf{m}}}} \right)}^T}{{\bf{U}}^{\bf{m}}}} \right)\\
		&- 2{\mathcal T}{ \times _1}{{\bf{T}}^T}{ \times _2}{\left( {{{\bf{I}}^{\bf{m}}}} \right)^T}{ \times _3}{\left( {{{\bf{U}}^{\bf{m}}}} \right)^T} + 2\alpha\left( {\mathcal A}-{\mathcal A}_0\right)+ 2\beta {\mathcal A} \\
		&+ 2{\lambda _1}\sum\limits_{i = 1,j = 1}^{|{\mathbb I}_n|} {{\bf{W}}_{i,j}^I\left( {{\mathcal A}{ \times _2}\left( {{{\bf{a}}_i}^T{{\bf{a}}_i}} \right) - {\mathcal A}{ \times _2}\left( {{{\bf{a}}_i}^T{{\bf{a}}_j}} \right)} \right)} \\
		&+ 2{\lambda _2}\sum\limits_{i = 1,j = 1}^{|{\mathbb U}_n|} {{\bf{W}}_{i,j}^U\left( {{\mathcal A}{ \times _3}\left( {{{\bf{b}}_i}^T{{\bf{b}}_i}} \right) - {\mathcal A}{ \times _3}\left( {{{ {{{\bf{b}}_i}} }^T}{{\bf{b}}_j}} \right)} \right)},
	\end{aligned}
\end{equation}
where 
${{\bf{a}}_i} = {\bf{I}}_{i,:}^m$, ${{\bf{a}}_j} = {\bf{I}}_{j,:}^m$,  ${{\bf{b}}_i} = {\bf{U}}_{i,:}^m$, and ${{\bf{b}}_j} = {\bf{U}}_{j,:}^m$. Therefore, the multiplicative updating rule~\cite{lee1999learning} of ${\mathcal A}$ is
\begin{equation}
\label{eq12.1}
{{\cal A}_{i,j,k}} = {{\cal A}_{i,j,k}} \frac{{{{\left( { {\cal H} + \alpha {{\cal A}_0} + {\lambda _1}{\cal Q} + {\lambda _2}{\cal P}} \right)}_{i,j,k}}}}{{{{\left( {{\cal G }} + (\alpha  + \beta ){\cal A} + {\lambda _1}{\cal U} + {\lambda _2}{\cal V} \right)}_{i,j,k}}}},
\end{equation}
where 
\begin{equation}
\begin{aligned}  
	&{\cal H} = {\cal T}{ \times _1}{{\bf{T}}^T}{ \times _2}{\left( {{{\bf{I}}^{\bf{m}}}} \right)^T}{ \times _3}{\left( {{{\bf{U}}^{\bf{m}}}} \right)^T},\\ 
	&{\cal G}={\cal A}{ \times _1}\left( {{{\bf{T}}^T}{\bf{T}}} \right){ \times _2}\left( {{{\left( {{{\bf{I}}^{\bf{m}}}} \right)}^T}{{\bf{I}}^{\bf{m}}}} \right){ \times _3}\left( {{{\left( {{{\bf{U}}^{\bf{m}}}} \right)}^T}{{\bf{U}}^{\bf{m}}}} \right),\\
	 	&{\cal Q}=\sum\limits_{i = 1,j = 1}^{|{\mathbb I}_n|} {{\bf{W}}_{i,j}^I\left( {{\cal A}{ \times _2}\left( {{{\bf{a}}_i}^T{{\bf{a}}_j}} \right)} \right)},\\
	 	 &{\cal P}=\sum\limits_{i = 1,j = 1}^{|{\mathbb U}_n|} {{\bf{W}}_{i,j}^U\left( {{\cal A}{ \times _3}\left( {{{\bf{b}}_i}^T{{\bf{b}}_j}} \right)} \right)},\\
	 	 &{U\cal }= \sum\limits_{i = 1,j = 1}^{|{\mathbb I}_n|} {{\bf{W}}_{i,j}^I\left( {{\cal A}{ \times _2}\left( {{{\bf{a}}_i}^T{{\bf{a}}_i}} \right)} \right)},\\
	 	 &{\cal V}=\sum\limits_{i = 1,j = 1}^{|{\mathbb U}_n|} {{\bf{W}}_{i,j}^U\left( {{\cal A}{ \times _3}\left( {{{\bf{b}}_i}^T{{\bf{b}}_i}} \right)} \right)}.
\end{aligned} 
\end{equation}
The details of algorithm are described in Algorithm~\ref{alg1}. Here, ${\mathcal A}$ is initialized by ${\mathcal A}_0+{\mathcal E}$, where ${\mathcal E}$ is a random small-disturbed tensor with mean 0. The convergence criterion is  that the  iterations  stop  when the  relative cost of the objective function is smaller than a predefined threshold $10^{-5}$. The proposed SUGAR-TC convergences after about $450$ iterations in the experiments.

\subsection{Anchor-Aware Tag Assignment}
After obtaining the completed tensor $\mathcal {A}$, we accumulate its entries along the user axis and image axis to acquire the desired tag-image association matrix ${\mathcal {A}}_{[3]}$ and tag-user association matrix ${\mathcal {A}}_{[2]}$, respectively. Then, we can employ the completed tags associated with anchor units to predict tags for non-anchor images. For one non-anchor image ${\bf x}_i$ ($i=1,2,\cdots, |{\mathbb{I}}_n|$) with an available user ${\bf u}_k$ ($k=1,2,\cdots, |{\mathbb{U}}_n|$), we use the weighted average of tags of $s$ nearest-neighbor anchor units to estimate its tag vector ${\bf y}_i$, as follows,
\begin{equation}
\label{eq12}
		{\bf{y}}_i \!=\!   \frac{\gamma {{\bf{I}}_{i,<\!i\!>}^m \!\left[\!\left({\bf{A}}_{[3]}\right)_{:,<\!i\!>}\right]^T \!+\!(1-\gamma) {{\bf{U}}_{k,<\!k\!>}^m\!\left[\!\left({\bf{A}}_{[2]}\right)_{:,<\!k\!>}\right]^T}}}{s},
\end{equation}
where $<\!i\!>$ is an index set of $s$ closest anchor-images of  ${\bf x}_i$, $<\!j\!>$ is an index set of users corresponding to these $s$ closest anchor-images, $\gamma$ is a parameter to control the degrees of tag-image and tag-user associations. Finally, we rank the elements of ${\bf y}_i$ based on the values in the descending order, and select top $10$  tags as the final tags of ${\bf x}_i$.

\section{Experiments}

\subsection{Experimental Settings}
We conduct experiments on a real-world social image dataset NUS-WIDE-128~\cite{tang2016generalized} to evaluate the performance of the proposed SUGAR-TC method. It is extended from the widely-used NUS-WIDE dataset~\cite{chua2009nus}, and contains $269,648$ images, $5,018$ user-provides tags crawled from the Internet, as well as the manually labeled ground-truth of 128 predefined concepts for evaluation. In this dataset, some user IDs are invalid or unavailable, thus we delete the corresponding images. Finally, we obtain $247,849$ images with $49,528$ user IDs and the information of user groups. 


In the experiments, we compare the proposed method with TRVSC~\cite{liu2010image}, LR~\cite{zhu2010image}, LSLR~\cite{li2016locality}, NMC~\cite{xu2017non}, MRTF~\cite{sang2011exploiting}, TTC1~\cite{tang2017tri} and TTC2~\cite{tang2017tri}. Besides, we also set a baseline method called Original Tagging (OT), where the labels used for evaluation are the user-provided tags crawled from Internet. To evaluate the performance, the widely-used F-score is adopted. For each concept, F-score is calculated as F-score$=\frac{2\times \text{Precision}\times \text{Recall}}{\text{Precision}+\text{Recall}}$. And then the average F-score over all the concepts is reported. For fair comparison, we use the CNN features~\cite{krizhevsky2012imagenet} and tune the hyper-parameters by using the grid search strategy for all methods in the experiments. The best results are reported for comparison. All methods in the experiments are implemented on a server with an 8-core 2.67 GHz CPU and 32 GB memory.

\begin{table*}[t!]
	\renewcommand{\arraystretch}{1.2}
	\centering
	\caption{Average F-scores of different methods for image retagging.}
	\vspace{-0mm}
	\label{tab_result}
	\begin{tabular}{c|ccccccccc}
		\hline
		Methods & OT& TRVSC~\cite{liu2010image} & LR~\cite{zhu2010image}& NMC~\cite{xu2017non} & LSLR~\cite{li2016locality}& MRTF~\cite{sang2011exploiting} & TTC1~\cite{tang2017tri}& TTC2~\cite{tang2017tri} & SUGAR-TC\\
		\hline
		Average F-scores & 0.379 & 0.390  & {0.410} & 0.417 & 0.435& 0.423& 0.447& 0.458 & 0.485\\
		\hline
	\end{tabular}
\end{table*}

%

\begin{figure*}[!t]
	\small{
		\centering	
			\includegraphics[scale=0.535]{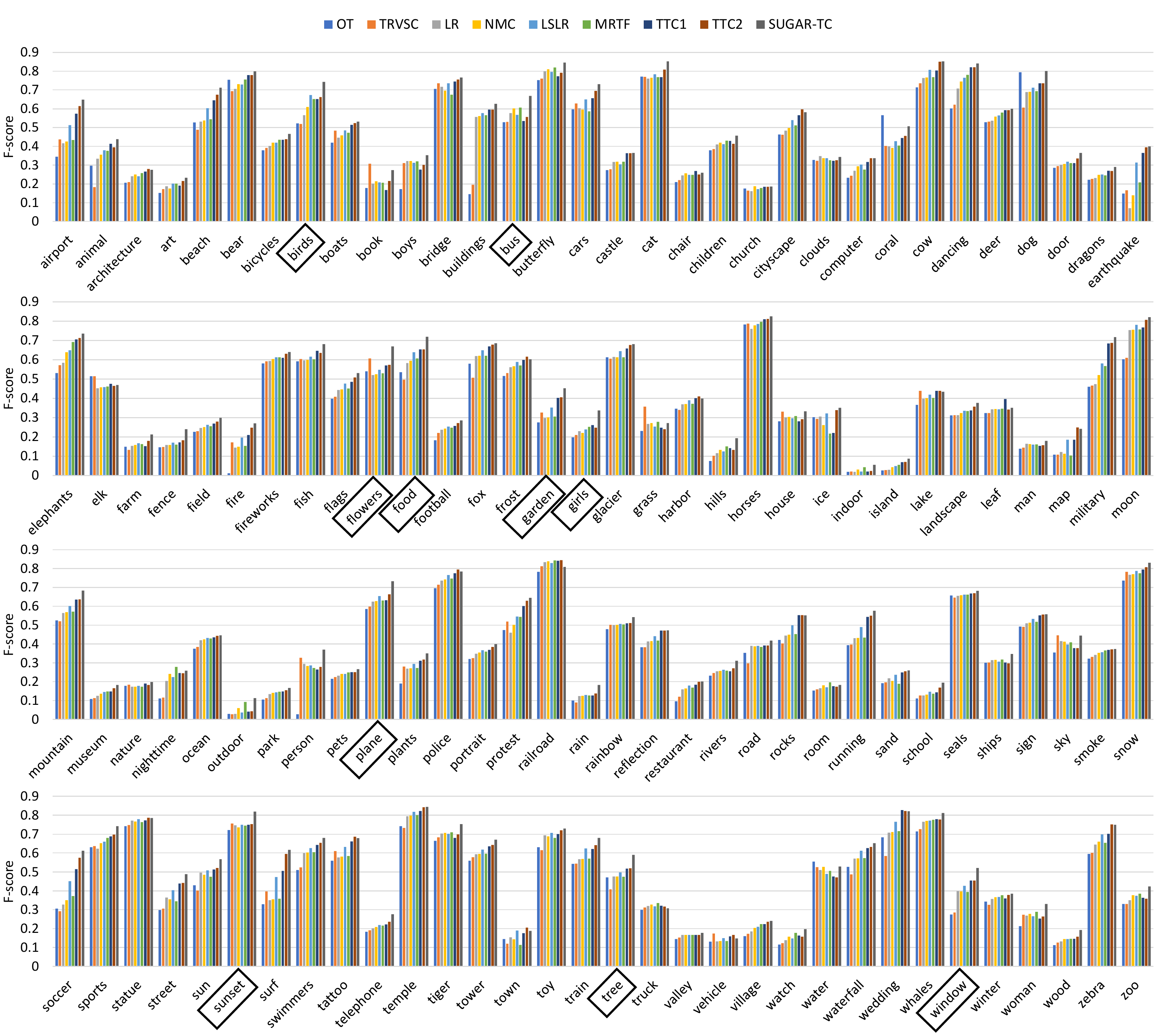}
			\vspace{-2mm}
		\caption{The comparisons of detailed F-scores of different methods. For those concepts denoted with the black bounding box, the proposed SUGAR-TC shows remarkable improvements. Best view in color.}
		\vspace{0mm}
		\label{fig_fscore}
	}
\end{figure*}


\subsection{Results and Analysis}
In this section, we conduct experiments to evaluate the effectiveness of the proposed SUGAR-TC. The compared results in terms of the average F-score are presented in Table~\ref{tab_result}. We find that the proposed SUGAR-TC achieves the highest F-score compared with the other related methods. 
MRTF, TTC1 and TTC2 utilize the inter- and intra- associations among images, tags and users, thus they perform better than TRVSC, LR, NMC and LSLR. In particular, since TTC1 and TTC2 address the super-sparse problem existing in the original tensor, they gain higher F-scores than MRTF that directly completes the entries in the original tensor. However, to a certain degree, both TTC1 and TTC2 break the latent inter-association among images, tags and users, when dividing the original tensor into several sub-tensors. In turn, the proposed SUGAR-TC did not break the inherent tag-image-user inter-association. Thus, the F-score obtained by SUGAR-TC is $0.485$, which is about $0.027$ higher than $0.458$ of TTC2.


\begin{figure}[t]
	\centering
	\includegraphics[scale=0.3400]{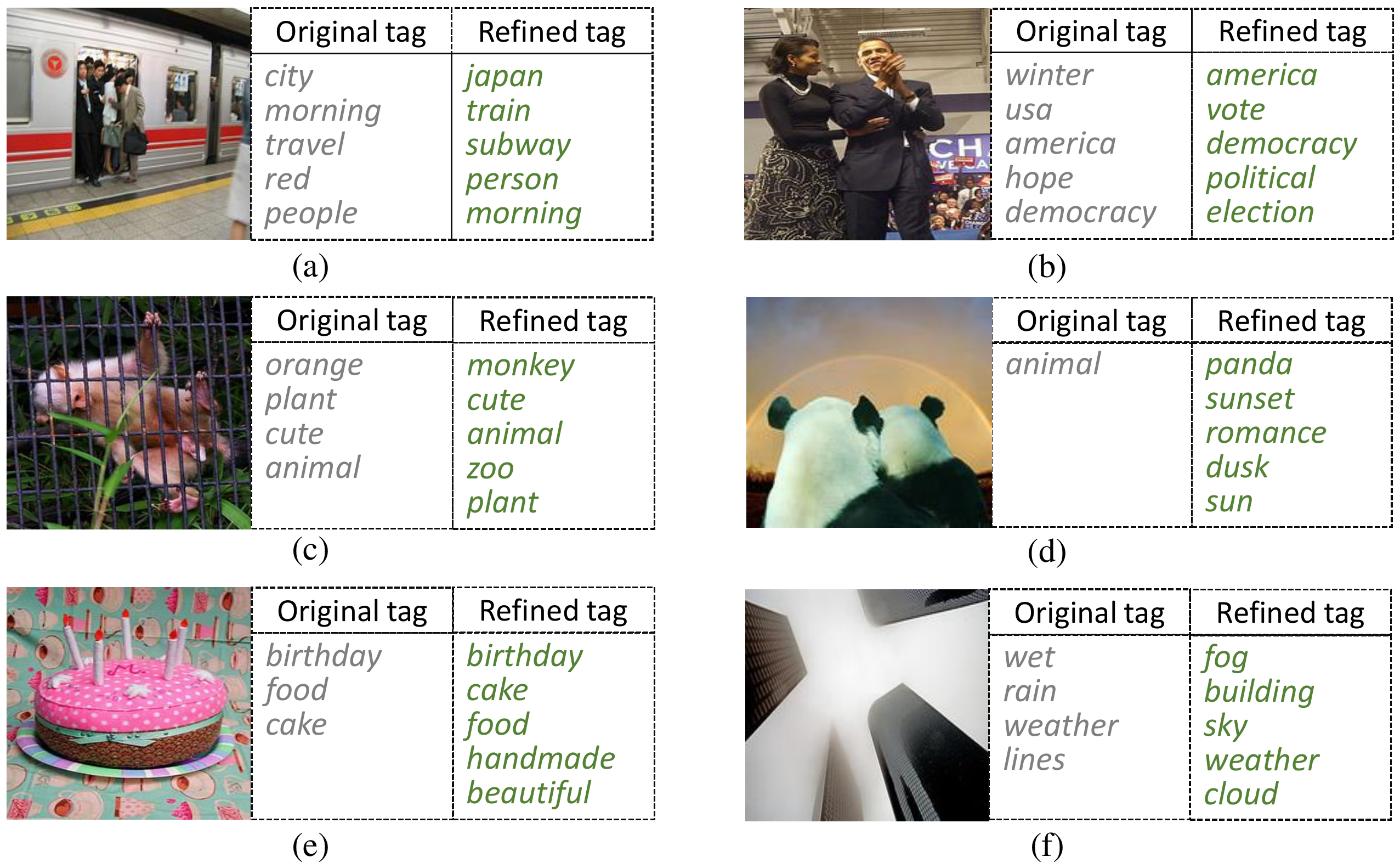}
	\vspace{-5mm}
	\caption{Some results of image retagging obtained by SUGAR-TC. }
	\label{fig_example}
\end{figure}

We further detail the F-scores obtained by different methods on all the 128 predefined concepts, as shown in Figure~\ref{fig_fscore}. We can see that SUGAR-TC achieves the highest F-scores on most of these concepts. We can also see that all image retagging methods improve the quality of almost all concepts compared with OT. By adding user information, MRTF, TTC1, TTC2 and SUGAR-TC perform better than LR, NMC, LSLR and TRVSC, especially for some summarized or complex tags (\textit{\textit{e.g.}}, ``military", ``nighttime",  and ``cityscape"). 
In particular, for the frequently-used tags (\textit{\textit{e.g.}}, ``bus", ``flowers", and ``sunset", etc) in Figure~\ref{fig_fscore}, SUGAR-TC shows remarkable improvement than TTC1 and TTC2, one reason may be that TTC1 and TTC2 break the tag-image-user inter-association for such common tags when decomposing the original tensor into several sub-tensors.

\begin{figure*}[!t]
	\small{
		\centering
		\subfigure[$C_i$]
		{
			\includegraphics[scale=0.24500]{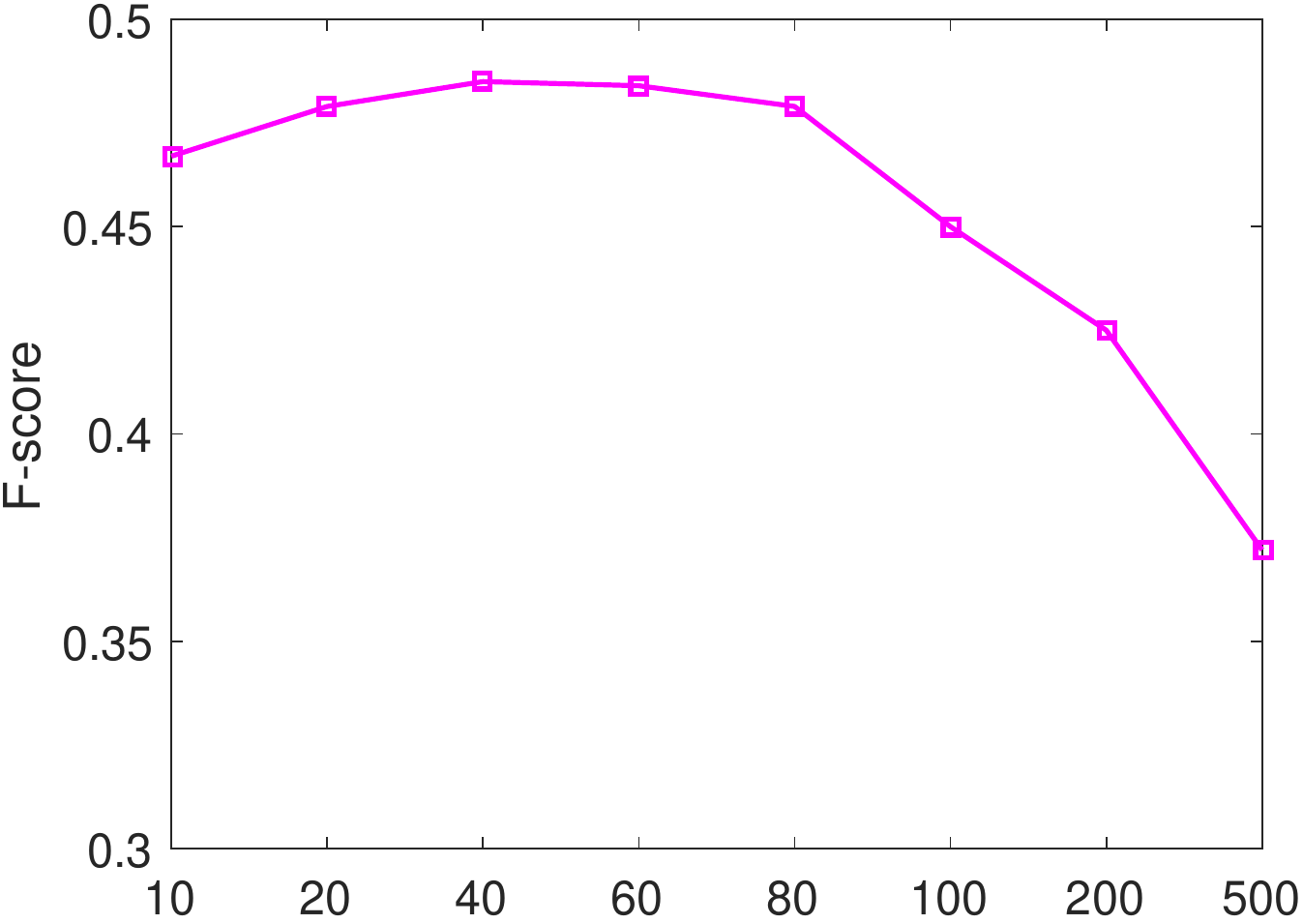}
			\label{fig5a}
		}
		\subfigure[$C_u$]
		{
			\includegraphics[scale=0.2400]{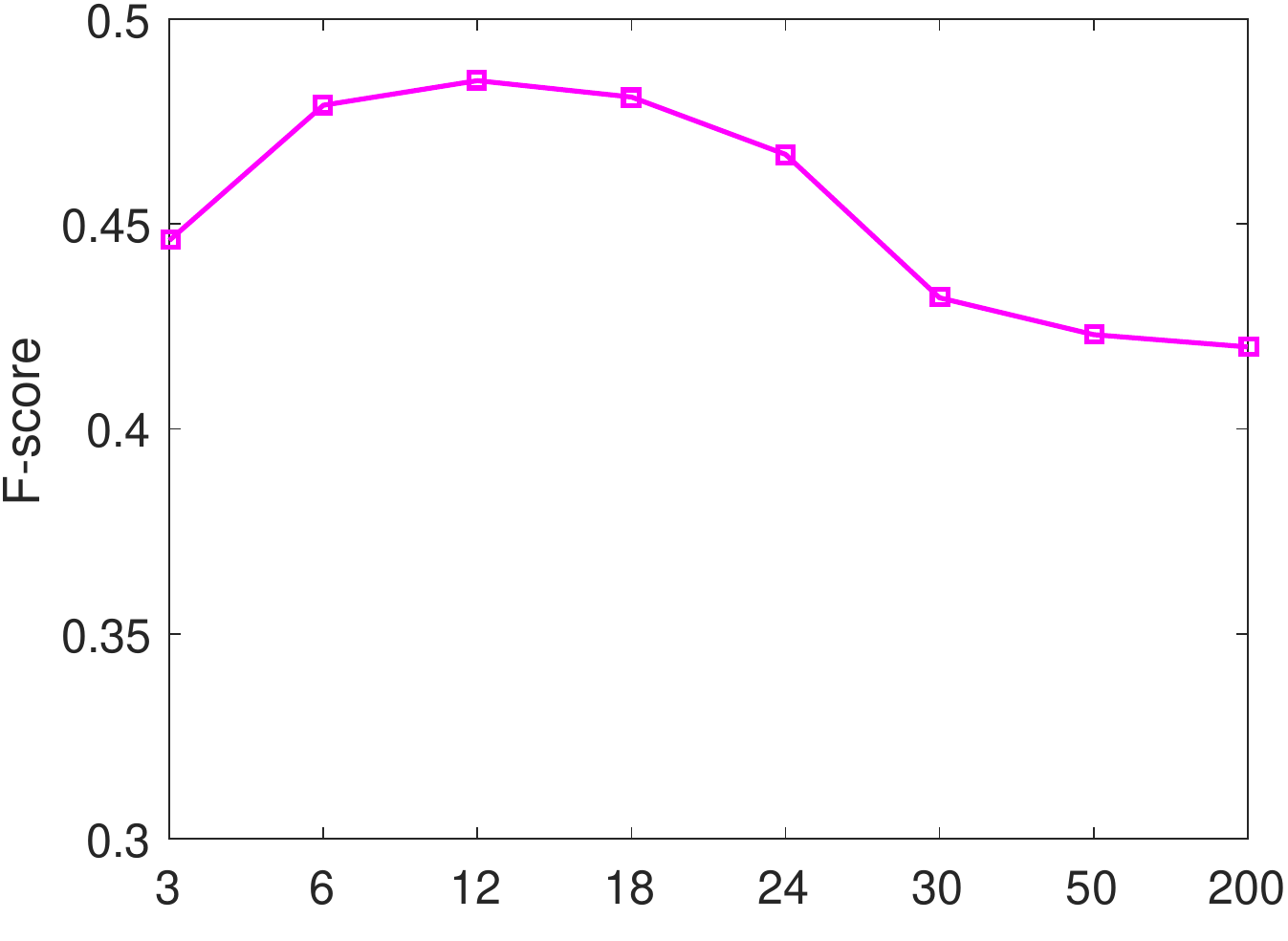}
			\label{fig5b}
		}
		\subfigure[$\alpha$ and $\beta$]
		{
			\includegraphics[scale=0.23500]{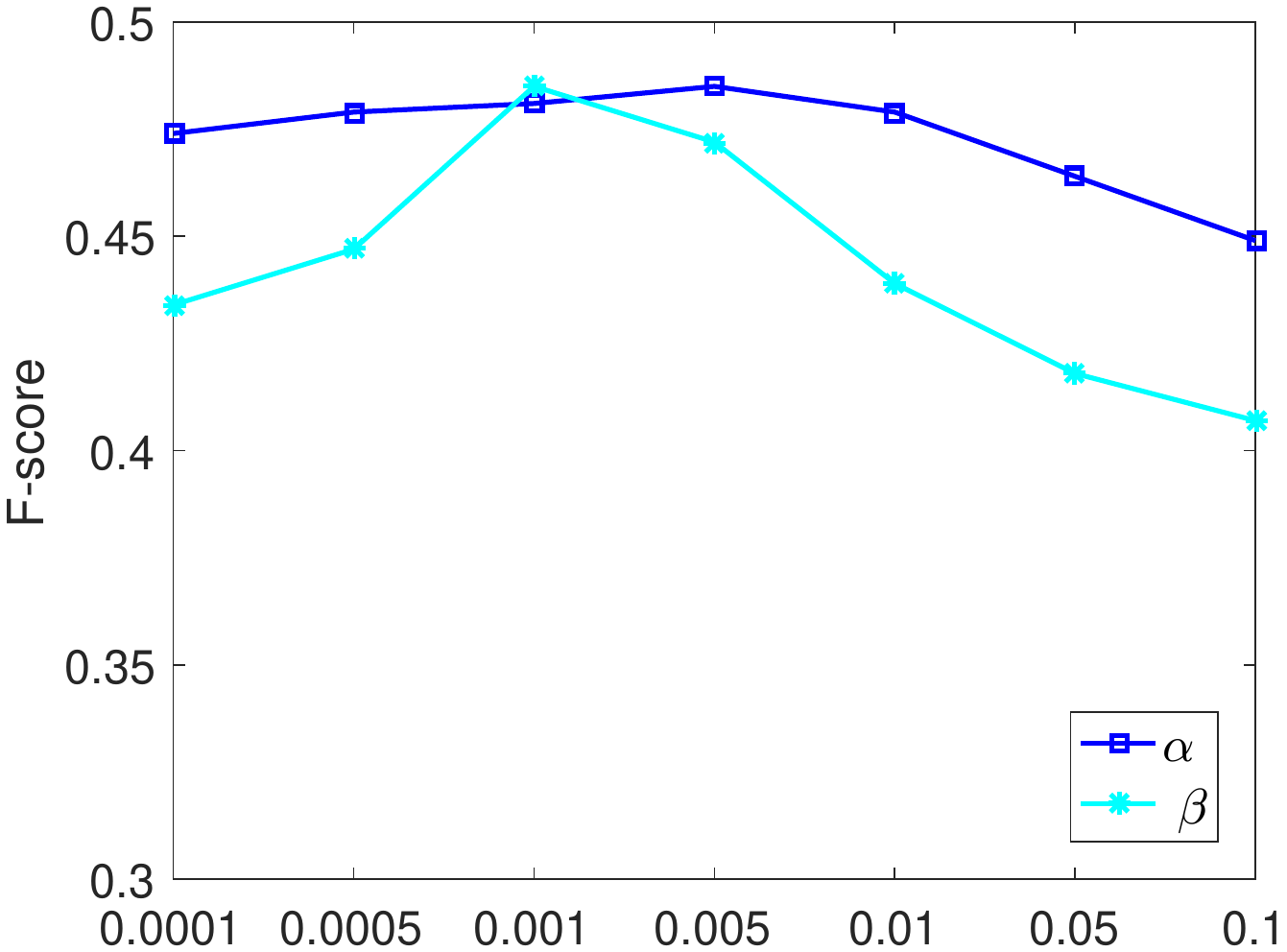}
			\label{fig5c}
		}
		\subfigure[$\lambda_1$ and $\lambda_2$]
		{
			\includegraphics[scale=0.24500]{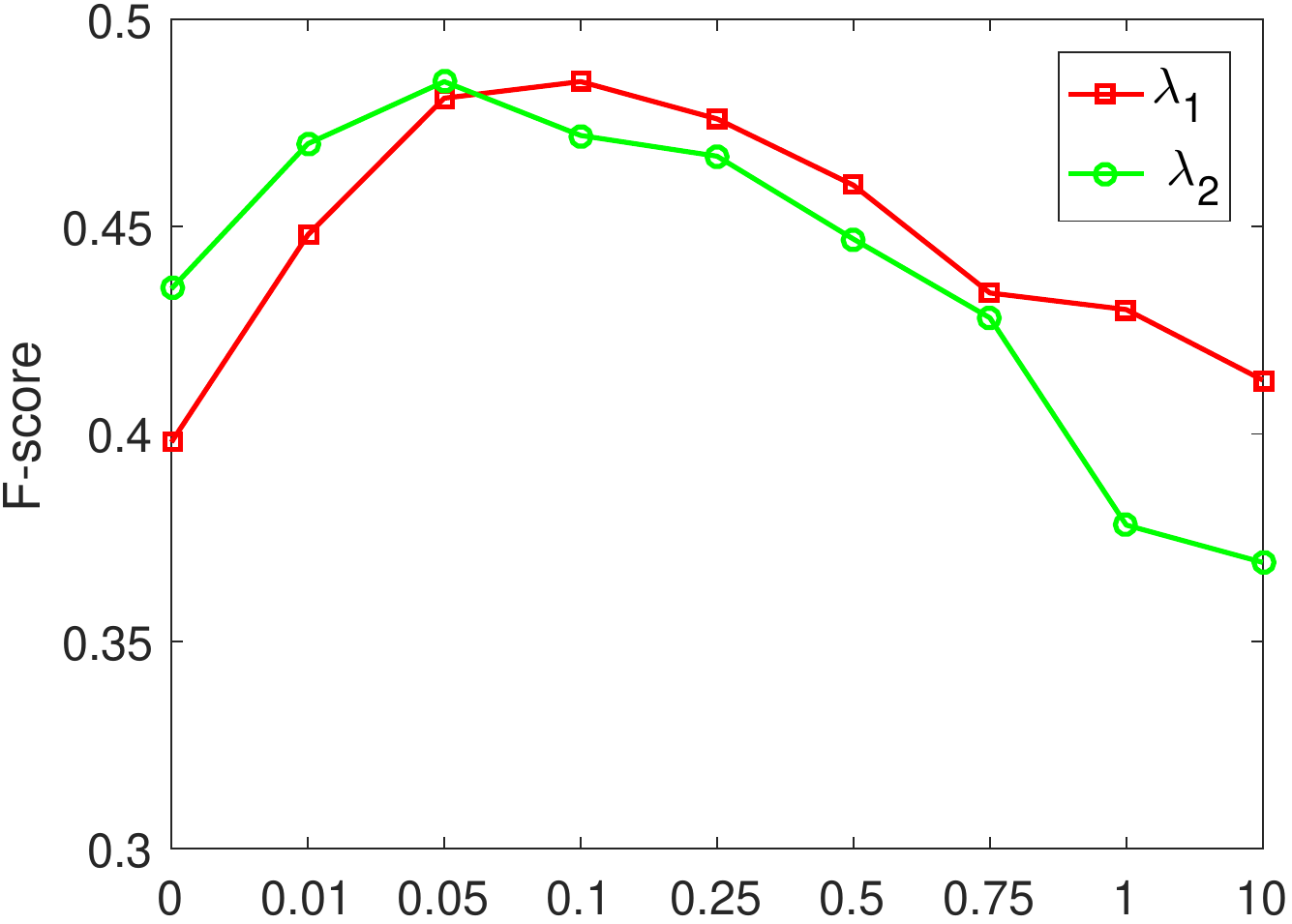}
			\label{fig5d}
		}
		\subfigure[$\gamma$]
		{
			\includegraphics[scale=0.24500]{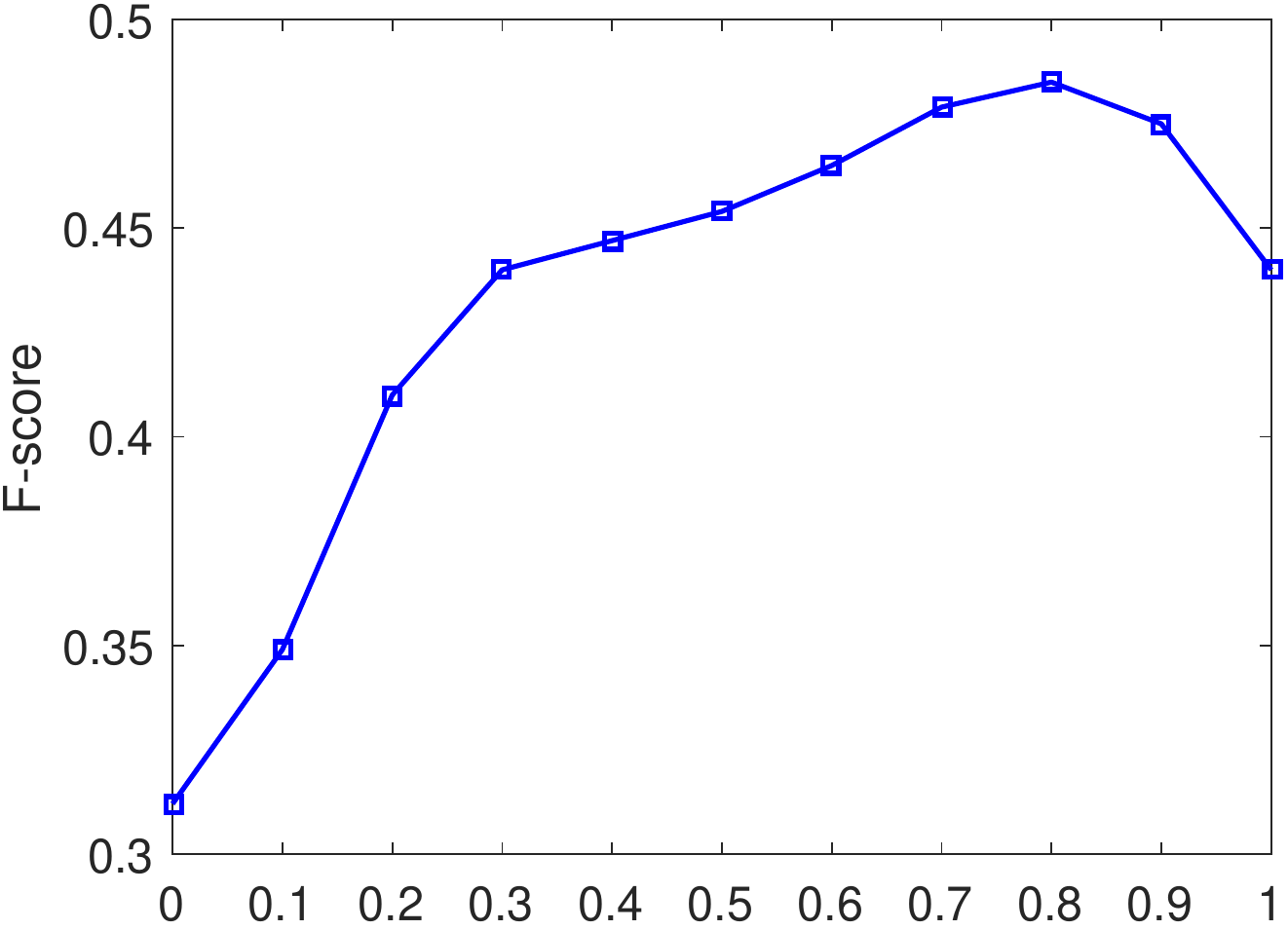}
			\label{fig5e}
		}
		
		
		\vspace{-2mm}
		\caption{The curves of F-scores obtained by varying one parameter with the other parameters fixed to be the default values.}
		\vspace{-0mm}
		\label{para}
	}
\end{figure*}

We also show some tag assignment results for some specific cases, as shown in Figure~\ref{fig_example}.
We can see that SUGAR-TC can effectively assign tags of images, even though there are some globally complex images (\textit{\textit{e.g.}}, Figure~\ref{fig_example}(c)) and locally abstract images (\textit{\textit{e.g.}}, Figure~\ref{fig_example}(d)), since SUGAR-TC can simultaneously mine the context information between users and images, and the semantic relation between images and tags. Besides, although some tags, \textit{i.e.}, geo-related tags, event tags, and time-related tags, are hardly inferred by only using visual information, SUGAR-TC considering extra user information can correctly infer them. For example, as shown in Figure~\ref{fig_example}(a), the train is a landmark of Japan, thus this image should be tagged with the geo-related tags ``Japan". It is very difficult to reveal the relation between the tag ``Japan" and this image by only mining the visual and tag information, but SUGAR-TC  can do it well based on the user background information. Similarly in Figure~\ref{fig_example}(b), this image shows the  Obama’s speaking of the United States presidential election, it is reasonable to assign the event tag ``election" to this image.


\subsection{Computational Cost Analysis}
\label{CCA}
To illustrate the efficiency of the proposed SUGAR-TC, we also
compare the computational cost of SUGAR-TC and the closely related
methods, including MRTF~\cite{sang2011exploiting}, TTC1~\cite{tang2017tri}, and TTC2~\cite{tang2017tri}. Table~\ref{time} lists the computing time
and memory cost of these methods in details. We can see that: 1) SUGAR-TC with total computing time 4.2h runs much faster than MRTF, TTC1 and TT2; 2) the memory cost of SUGAR-TC is more than TTC1 and TTC2, but less than MRTF. By jointly considering the accuracy, computing time and memory cost, SUGAR-TC is more applicable for image retagging.

\begin{table}[!t]
	\centering
	\caption{Detailed comparisons of computational time (hour) and memory (GB) on NUS-WIDE-128. `` / " denotes ``time/memory".}
	\vspace{-1mm}
	\label{time}
	\begin{tabular}{l|c|c|c|c}
		\hline 
		Method & MRTF~\cite{sang2011exploiting} & TTC1~\cite{tang2017tri} & TTC2~\cite{tang2017tri} & SUGAR-TC  \\
		\hline
		Clustering &-	&4.3/8.6&	4.3/8.6&	2.1/5.5\\
		Adjacency matrix & 1.6/1.8 & 	1.6/1.8&	1.6/1.8&	0.7/0.07\\
		Tensor completion & 12.3/10.3&	5.7/6.6&	29.6/8.2&	1.2/10.1\\
		Tag assignment &-&	-	&-&	0.2/0.08\\
		\hline 
		Total time &13.9	&11.6&	35.5&	4.2\\
		\hline 
	\end{tabular}
\vspace{-0mm}
\end{table}

\subsection{Parameter Sensitivity Analysis}
\label{PSA}
In the proposed SUGAR-TC method, there are several hyper-parameters to be set in advance. Following \cite{tang2017tri}, the weight coefficients $a_1$ and $a_2$ in Eq.~\eqref{eq4.5} are set to $0.9$ and $0.1$ respectively. The radius parameter $\delta$ in Eq.~\eqref{eq1} is set to $2.5$. The number of anchor units is set as $m_c=10$ for each co-cluster center. In the following, we will discuss the sensitiveness of other parameters. 

For the number of image centers (denoted by $C_i$) and the number of user clusters (denoted by $C_u$), we tune them by $C_i \in \{$10, 20, 40, 60, 80, 100, 200, 500$\}$ and $C_u \in \{$3, 6, 12, 18, 24, 30, 50, 200$\}$, respectively. The corresponding F-scores are shown in Figure~\ref{fig5a} and Figure~\ref{fig5b}, respectively. From the results, we can see that the best result is obtained with $C_i=40$ and $C_u=12$. When these two parameters are set within $[20, 80]$ and $[6, 18]$ respectively, the performance changes slightly. In the experiments, the default number of image clusters and user clusters are set to 40 and 12, respectively.

For $\alpha$, $\beta$, $\lambda_1$, $\lambda_2$, and $\gamma$, we tune them by $\alpha=\{$0.0001, 0.0005, 0.001, 0.005, 0.01, 0.05, 0.1$\}$, $\beta=\{$0.0001, 0.0005, 0.001, 0.005, 0.01, 0.05, 0.1$\}$, $\lambda_1=\{$0, 0.01, 0.05, 0.1, 0.25, 0.75, 1, 10$\}$, $\lambda_2=\{$0, 0.01, 0.05, 0.1, 0.25, 0.75, 1, 10$\}$, and  $\gamma=\{$0, 0.1, 0.2, 0.3, 0.4, 0.5, 0.6, 0.7, 0.8, 0.9, 1$\}$. The corresponding results are shown in Figure~\ref{para}(c-e). SUGAR-TC achieves the best result with $\alpha=0.005$, $\beta=0.001$, $\lambda_1=0.1$, $\lambda_2=0.05$, and $\gamma=0.8$, which are set as the default values of these parameters in the experiments.

\subsection{Application: Image Retrieval}
In this work, for each image, image retagging task can not only add and remove
tags, but also re-rank the completed tag list by assigning tags with different confidence scores. For each image, we can obtain a tag ranking list, which can improve the performance of tag-based image retrieval. Therefore, we conduct experiment of tag-based image retrieval to illustrate the effectiveness of SUGAR by comparing with other related methods, including TRVSC~\cite{liu2010image}, LR~\cite{zhu2010image}, NMC~\cite{xu2017non}, LSLR~\cite{li2016locality}, MRTF~\cite{sang2011exploiting}, TTC1~\cite{tang2017tri} and TTC2~\cite{tang2017tri}.

 \begin{figure}[t]
	\centering
	\includegraphics[scale=0.522]{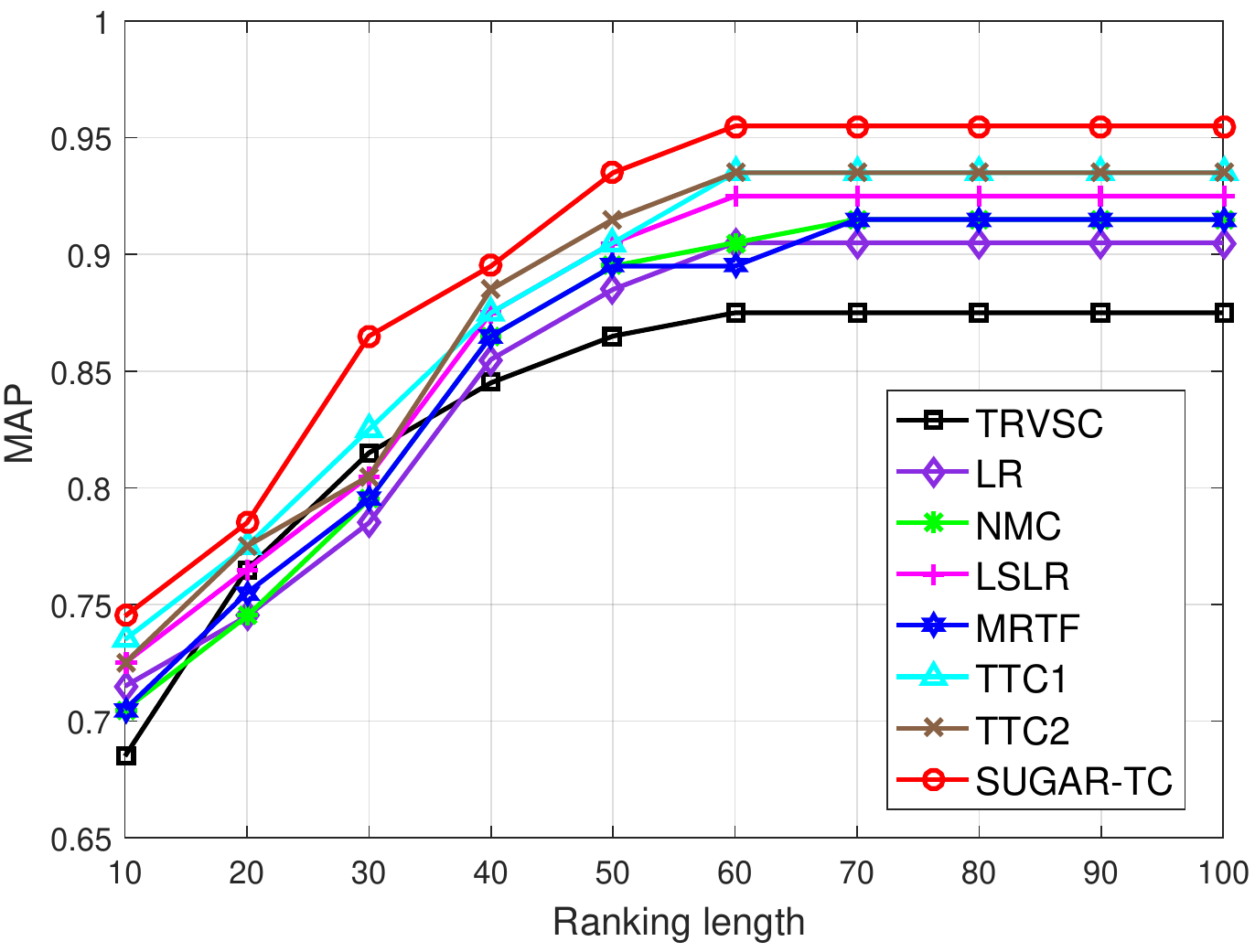}
	\vspace{-2mm}
	\caption{MAP of different methods for tag-based image retrieval.}
	\label{fig_MAP}
	\vspace{-2mm}
\end{figure}


Following the experimental setting in~\cite{liu2010image}, we perform tag-based social image retrieval with ten queries on NUS-WIDE-128, \textit{i.e.}, \textit{birds}, \textit{building}, \textit{butterfly}, \textit{dog}, \textit{fish}, \textit{flowers}, \textit{horses}, \textit{plane}, \textit{sunset} and \textit{zoo}. Average Precision (AP) is used as the evaluation metric. Specifically, given a ranked list with length $T$, AP is defined as $AP = \frac{1}{R}\sum\nolimits_{r = 1}^T {p(r) \cdot \chi (r)} $. Here, $R$ is the number of relevant images in this ranked list, $p(r)$ is the precision at cut-off top $r$, $\chi (r)$ is an indicator function:  $\chi (r)=1$ if the $r$-th image is relevant with respect to the ground-truth concept, otherwise $\chi (r)=0$. Finally, we use MAP over all queries to evaluate the overall performance. Figure~\ref{fig_MAP} shows MAP values with different ranking length T. We can see that
the proposed SUGAR-TC outperforms the other four image retagging methods.  This demonstrates  SUGAR-TC is more effective than other related image reatgging methods in terms of tag-based social image retrieval.


\section{Conclusions}
In this work, we propose a novel Social anchor-Unit GrAph Regularized Tensor Completion (SUGAR-TC) method to efficiently refine the tags of social images, regardless of the data scale. First, we utilize the co-clustering algorithm to obtain the representative anchor units (anchor image and anchor user) across image and user domains rather than traditional anchors in a single domain, and then construct an anchor-unit graph with multiple intra/inter adjacency edges. Second, we present a SUGAR tensor completion to refine tags of anchor images. Finally, we efficiently assign high-quality tags to all non-anchor images by leveraging the potential
relationships between non-anchor units and anchor units. Experimental
results on a real-world social image database demonstrate the effectiveness and efficiency of SUGAR-TC compared with the state-of-the-art methods.



\bibliographystyle{IEEEtran}
\bibliography{IEEEabrv,sample-bibliography-short}

\end{document}